\documentclass[letterpaper,10pt,conference]{ieeeconf}
\IEEEoverridecommandlockouts
\usepackage{graphicx}
\usepackage{mathptmx}
\usepackage{amsmath,amssymb,mathtools}
\usepackage{booktabs,tabularx,array}
\usepackage{microtype}
\usepackage{url}
\usepackage{tikz} 
\DeclareRobustCommand{\site}[1]{\tikz[baseline=(s.base)]\node[circle,draw,line width=0.4pt,inner sep=0pt,minimum size=7.5pt,font=\tiny\bfseries] (s) {#1};}
\graphicspath{{figures/}}
\newcommand{\figscale}{0.88}
\newcommand{\floatgap}{\vspace{-1.5mm}}      
\newcommand{\floatgapwide}{\vspace{-2mm}}    
\title{\LARGE\bf Decodable Targets but Incorrect Actions: Improving Sequencing Generalization of a VLA Policy with Sharpness-Aware Fine-Tuning}

\author{Yeongseok An and Junhyoung Ha$^{*}$%
\thanks{This work has been submitted to the IEEE for possible publication. Copyright may be transferred without notice, after which this version may no longer be accessible.}%
\thanks{The authors are with the Department of Mechanical Engineering, Ulsan National Institute of Science and Technology (UNIST), Ulsan, Republic of Korea. {\tt\small \{cieclr, j.ha\}@unist.ac.kr}}%
\thanks{$^{*}$Corresponding author.}%
}

\begin{document}
\maketitle
\begin{abstract}
A vision-language-action policy can select the wrong next object even when the instructed target remains decodable from its backbone and VL-adapter states. We study this mismatch in GR00T N1.7 on three-cube sequencing, holding out red--green--blue (RGB) while fine-tuning five other orderings; the only red-first training ordering continues to blue, whereas RGB requires green. Under AdamW at 20,000 updates, held-out RGB succeeds in 65\% of rollouts against an 86.6\% seen-ordering mean, with wrong-blue second placements in 20\% of rollouts. In the 10,000-update AdamW checkpoint used for representation analyses, the VL adapter shows no detectable failure-associated deficit in instruction-conditioned geometric separation, whereas a deficit is present throughout the DiT action-token pathway; where it first arises is not identified. At a matched forward--backward-pass budget and without RGB demonstrations, Sharpness-Aware Minimization (SAM) reaches 81\% RGB success versus 65\% for AdamW 20k, with similar seen-ordering means (86.8\% versus 86.6\%), fewer wrong-blue second placements (4\% versus 20\%), and a reduced red-first asymmetry on disjoint layouts. On a physical SO-101 with separately trained policies, AdamW 20k follows the instructed order in 18/30 RGB versus 26/30 RBG trials and SAM in 24/30 versus 29/30, with blue-second placements the most frequent RGB failure for both. On the same post-red observations, the separation gap between AdamW-defined success and failure cohorts is no longer detectable under SAM. The findings distinguish upstream target decodability from downstream geometric separation and are consistent with a continuation-shortcut account in this single-seed setting, without establishing a causal or optimizer mechanism.
\end{abstract}

\section{Introduction}
Vision-language-action (VLA) models map camera observations, robot state, and language instructions to robot actions. RT-2, OpenVLA, Octo, and GR00T demonstrate the promise of transferring pretrained knowledge across manipulation settings~\cite{brohan2023rt2,kim2024openvla,octo2024,nvidia2025gr00tn1}. Yet performing familiar actions in a new instructed order remains difficult~\cite{jiang2022vima,peng2026acvla}. A wrong object choice alone does not reveal whether target information is absent from the policy's internal representations or remains present without leading to the instructed behavior.

We study this problem through three-cube sequencing: a robot must place red, green, and blue cubes into a white box in the order specified by a language instruction. A task denotes one instructed ordering, and success requires all three placements in that order. We abbreviate the colors as R, G, and B; RGB therefore means red first, green second, and blue third. The five seen orderings---GRB, BGR, BRG, GBR, and RBG---are included in fine-tuning, while RGB is held out. Every object and individual pick-and-place skill is thus familiar, but the required RGB sequence is new.

The key decision comes immediately after red is placed in the box, which we call the post-red state. Both RGB and RBG begin with red, but only RBG appears in training: its next target is blue, whereas RGB requires green. We call the demonstrated red-to-blue transition a familiar continuation. Selecting blue under RGB would follow this continuation instead of the instruction. This motivates a continuation-shortcut account, drawing on work on causal confusion and shortcut learning~\cite{dehaan2019causal,geirhos2020shortcut}.

\begin{figure}[t]
\centering
\includegraphics[width=\figscale\columnwidth]{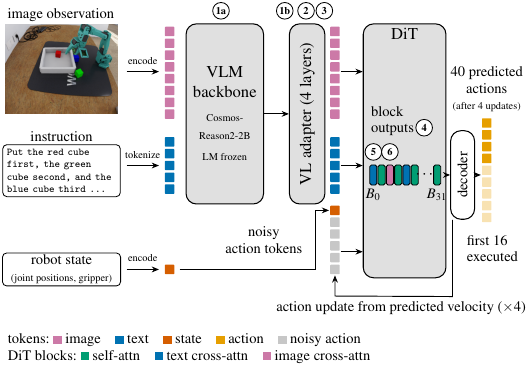}
\floatgap
\caption{Analyzed GR00T N1.7 pipeline and measurement sites. Backbone image tokens (64 per camera, external and wrist) and text tokens pass through the four-layer VL adapter and condition the DiT action tokens by cross-attention; the analyzed action-token states downstream of this adapter-to-DiT boundary form the action pathway. At each of four flow-matching steps the decoder's predicted velocity updates the action tokens; the first 16 of the 40 predicted actions are executed and analyzed. Numbered sites mark where analyses read or intervene: \site{1} target probe, \site{2} attention mass, \site{3} adapter $D_{\mathrm{rel}}$, \site{4} DiT block-output $D_{\mathrm{rel}}$, \site{5} cross-attention routing ($B_{8,28}$ text, $B_{10,26}$ image), \site{6} activation replacement ($B_{10,18,22,26}$).}
\label{fig:architecture}
\end{figure}

Figure~\ref{fig:architecture} introduces the processing stages of our case-study model, GR00T N1.7~\cite{nvidia2025gr00tn1,nvidia2026n17}: a vision-language backbone encodes the instruction and the external and wrist camera images; a vision-language (VL) adapter turns these tokens into context that, with the robot state, conditions a diffusion transformer (DiT) refining a noisy action sequence. The sequence's intermediate vectors are action tokens (action queries when they attend to the context). The policy predicts 40 actions and executes the first 16 before observing again.

For each saved post-red observation, we compare RGB and RBG instructions with images, robot state, and initial action noise fixed. A linear probe predicts target color from an internal representation; geometric separation measures how much that representation changes between instructions; activation replacement tests whether substituting RBG representations into the RGB computation changes predicted actions. In the primary analysis of the 10,000-update AdamW baseline, targets remain decodable from the upstream backbone and VL adapter, while failed rollouts show smaller instruction-conditioned separation in downstream DiT action tokens, and replacing later DiT states has stronger effects on decoded actions.

We then test whether the fine-tuning optimizer matters. Sharpness-Aware Minimization (SAM)~\cite{foret2021sam} penalizes high loss near the current weights without adding demonstrations or sequencing supervision; each SAM update uses two forward--backward passes, so SAM at 10,000 updates is matched with AdamW at 20,000. At perturbation radius $\rho=0.05$, SAM improves held-out RGB performance and reduces the red-first asymmetry on disjoint layouts---cube arrangements absent from the demonstrations---while matching the AdamW 20k seen-task mean. We also evaluate separately fine-tuned policies on a physical SO-101.

Our contributions, in a single-model, single-seed case study, are: (1) a sequence-recombination test contrasting the instructed next target with a demonstrated continuation; (2) an analysis distinguishing target decodability, instruction-conditioned separation, attention routing, and activation-replacement effects; and (3) a fine-tuning intervention that, at $\rho=0.05$, improves held-out RGB performance on familiar and disjoint layouts, with controls delimiting its interpretation.

\section{Related Work}
We connect work on sequence recombination, decodability versus behavioral influence, and sharpness-aware optimization.
\subsection{Generalist and Compositional Robot Learning}
Generalist policies combine heterogeneous robot data~\cite{openx2023} and pretrained vision-language representations~\cite{brohan2023rt2,kim2024openvla,octo2024}. GR00T and $\pi_0$ condition generative actions on multimodal context~\cite{nvidia2025gr00tn1,black2025pi0}, building on Diffusion Policy, Flow Matching, and DiT~\cite{chi2023diffusionpolicy,lipman2023flowmatching,peebles2023dit}. Recombination benchmarks~\cite{lake2018systematicity,hupkes2020compositionality,ruis2020gscan} and manipulation methods including VIMA, CLIPort, ProgramPort, and AC-VLA~\cite{jiang2022vima,shridhar2022cliport,wang2023programport,peng2026acvla} motivate our test of an instruction--continuation conflict.

\subsection{Shortcuts, Probing, and Interventions}
Training correlations can mislead imitation under distribution shift~\cite{dehaan2019causal,geirhos2020shortcut}. Probing measures decodability rather than use~\cite{belinkov2022probing,elazar2021amnesic}; attention describes routing, not causality~\cite{jain2019attention}. Activation replacement tests output influence~\cite{meng2022rome}, subject to intervention and metric design~\cite{zhang2024patching}. Recent VLA work combines probing and activation injection across models to characterize language dependence~\cite{grant2026features}. Our question is narrower and controlled: when the instructed next action after a shared prefix conflicts with the only continuation observed in training, how do the policy's choice and its internal representations differ? We answer it with matched-observation RGB/RBG comparisons in one policy, and test whether an optimizer change alters the outcome.

\subsection{Sharpness-Aware Optimization}
Sensitivity of loss to weight perturbations motivates generalization studies~\cite{keskar2017large} and SAM's bounded-perturbation objective~\cite{foret2021sam}. ASAM highlights sensitivity to parameter scaling~\cite{kwon2021asam}, while SAM has also benefited language modeling~\cite{bahri2022sam}.

\section{Experimental Setup and Methods}
We describe the task, training procedure, and behavioral evaluations, then define the saved observations used for representation analysis.
\subsection{Task, Demonstrations, and Layouts}
A robot places red, green, and blue cubes into a white box. Instructions use the fixed template ``Put the \{$c_1$\} cube first, the \{$c_2$\} cube second, and the \{$c_3$\} cube third into the white box.'' Fine-tuning uses GRB, BGR, BRG, GBR, and RBG; RGB is excluded. The wording is unchanged at evaluation, so this tests sequence recombination rather than paraphrase generalization.

A layout specifies the cubes' initial positions and orientations. We collect 20 teleoperated demonstrations per seen ordering (100 total), using the same 20 layouts across orderings. Each fixes planar position and yaw at $z=0.04\,\mathrm{m}$. Familiar-layout evaluation reuses these arrangements: RGB is out of distribution (OOD) in sequence, but not in initial layout. A separate behavioral control evaluates RGB and RBG on 20 additional layouts absent from the demonstration table. All mechanistic analyses use familiar-layout snapshots.

The seen next targets are $\{G,B,R\}$ initially, $\{R,B\}$ after G, $\{G,R\}$ after B, and B only after R; thus RGB requires the unsupported $R\!\rightarrow\!G$ continuation whereas RBG supplies $R\!\rightarrow\!B$. This regularity does not imply identical post-red pixels or simulator states across trajectories.

Evaluation uses the Isaac Lab SO-101 ordered-cube environment, in the simulation framework lineage of Orbit~\cite{mittal2023orbit}. Inputs comprise an external/front camera, wrist camera, robot state, and instruction; each camera contributes 64 image tokens.

\subsection{Policy and Fine-Tuning}
We use GR00T-N1.7-3B with a Cosmos-Reason2-2B backbone based on Qwen3-VL~\cite{nvidia2026n17}. The analyzed revision\footnote{Isaac-GR00T commit \texttt{3df8b382\allowbreak{}5d67f755\allowbreak{}e6914144\allowbreak{}6f4315f2\allowbreak{}81b9b7e6}; no architectural modifications were made.} exposes four VL-adapter self-attention layers and 32 DiT blocks, $B_0,\ldots,B_{31}$. Each policy query predicts a chunk of 40 consecutive actions; the first $K=16$ are executed and analyzed.

The AdamW~\cite{loshchilov2019adamw} runs use 10,000 or 20,000 optimizer updates, learning rate $10^{-4}$, weight decay $10^{-5}$, warmup ratio 0.05, global batch-size configuration 32, gradient accumulation 4 (128 samples per optimizer update), and seed 42. The language model is frozen; the visual module, projector, and action model are trainable. State dropout is 0.2 and relative-action training is enabled.

SAM uses perturbation radius $\rho=0.05$, 10,000 updates, one GPU, and AdamW as its base optimizer; data, flow-matching loss, architecture, and trainable components are unchanged. For the full accumulated-batch gradient $g$ over all trainable parameters, the perturbation is $e=\rho\, g/(\lVert g\rVert_2+10^{-12})$. SAM replays the same microbatches at $w+e$ with saved random-number-generator states, restores $w$, and updates using the second-pass gradient. Each update thus uses two forward--backward passes. No RGB demonstrations, target labels, phase annotations, or auxiliary sequencing losses are added. Additional $\rho=0.01$ and $\rho=0.1$ runs differ only in $\rho$.

SAM 10k and AdamW 20k match forward--backward passes, with approximate elapsed spans of 11\,h\,26\,min and 11\,h\,42\,min.\footnote{Estimated from run-output-directory birth to final-checkpoint-directory birth; these filesystem intervals are not instrumented training runtimes.} This pair defines the primary behavioral comparison on six familiar-layout orderings. AdamW 10k matches SAM's optimizer updates and supplies the primary representation analyses and disjoint-layout control. Unless specified, AdamW and SAM denote the 10k checkpoints.

\subsection{Behavioral Evaluation and Statistics}
Before each rollout, 45 control steps drive the robot to the demonstration frame-0 mean configuration $[-2.4,-93.3,66.5,77.6,-6.4,25.3]$ in the LeRobot joint-coordinate convention. The first 16 of the 40 predicted actions execute before the next query. Each ordering uses $20$ layouts $\times5$ repeats, or 100 rollouts; the disjoint control uses the same protocol per ordering and checkpoint. Repeats share initialization and homing and capture stochastic policy variation, not new layouts.

The primary score is environment-defined full-sequence success. We report percentile 95\% confidence intervals (CIs) from 100,000 layout-cluster bootstrap replicates, retaining all five repeats of each sampled layout. Checkpoint comparisons pair layout identities and use exact two-sided layout-level sign-flip/randomization tests. The five individual seen-task tests use Holm correction; other reported $p$-values are unadjusted. The same procedures apply to the 20k and disjoint controls. Success-rate differences are reported in percentage points (pp). Seen performance is the mean across five orderings, $\bar S_{\mathrm{seen}}$; the generalization gap is $G=\bar S_{\mathrm{seen}}-S_{\mathrm{RGB}}$.

\paragraph{Real-robot color-order evaluation}
For the physical SO-101, AdamW 20k and SAM 10k ($\rho=0.05$) are separately fine-tuned on 20 real-robot demonstrations per seen ordering (100 total), with the same five-order RGB holdout. Evaluation uses external and wrist cameras and the same instruction template. We evaluate 30 scored trials per ordering (RGB, RBG) and checkpoint with manually set, unmatched cube placements. Grasp failures trigger an immediate restart of the entire trial and are excluded from the color-order score; color-order errors count as failures and are categorized by the executed order as a wrong first target or a wrong second target after a correct first placement. Comparisons are unpaired, using Wilson 95\% CIs and two-sided Fisher exact tests.

\subsection{Failure Audit and Snapshot Cohorts}
To distinguish target-selection errors from execution failures, evidence is ranked as box entry, lift, then approach; the same post-hoc classifier audits every checkpoint (approach distance $0.10\,\mathrm{m}$, gripper-close threshold 30 for at least three steps, lift height $0.075\,\mathrm{m}$) and does not replace the environment success flag. Inferred wrong targets without entry remain separate from observed errors.

For the primary representation analyses, a separate one-rollout-per-layout collection saves 19 post-red snapshots (observations immediately after the first red placement; one layout does not reach this state), independent of the 100-rollout evaluation. AdamW 10k outcomes define success ($n=12$) and failure ($n=7$) cohorts; strict entry/lift evidence defines correct-green ($n=12$) and wrong-blue ($n=5$) cohorts, excluding two unresolved cases. Labels stay fixed when the same observations are evaluated under SAM. The AdamW 20k replication uses a separate 20-state cohort from its familiar-layout evaluation (first rollout per layout, in repeat order, to reach post-red; success/failure 16/4, strict 17/3), never pooled with the 10k states.

\subsection{Target Probing and VL-Adapter Attention}
A multinomial logistic-regression probe trained only on seen-task states~\cite{belinkov2022probing,elazar2021amnesic} predicts red, green, or blue from fixed Qwen and VL-layer representations (Fig.~\ref{fig:architecture}, \site{1}), reading the contextualized token(s) of ``second'' (mean-pooled if split); color-token and text-span pools serve as controls. RGB and counterfactual RBG are evaluation-only; their green/blue second targets form a balanced binary evaluation (baseline 50\%) of decodability, not progress-aware causal use.

For DiT (\site{4}), separate three-class linear probes (class-balanced logistic regression, $C=1$, per-fold standardization) use mean-pooled first-16 block-output tokens at $B_0$ and $B_{31}$, $t=3$ (the last of four denoising steps), from the AdamW checkpoint. Training uses 80 snapshot--instruction pairs (branches) from 20 after-green and 20 after-blue snapshots with verified first placements (red/green/blue: 40/20/20); validation leaves out each layout across both prefixes. Evaluation uses only the 19 post-red RGB/RBG pairs (no shared trajectories or evaluation-based tuning) with 20,000 paired-layout bootstrap replicates; CIs exclude probe-refitting uncertainty. The same protocol is evaluated with the AdamW 20k checkpoint, applied at feature extraction to the saved after-green/after-blue seen-task snapshots (80 branches, 40/20/20) and to the 20 post-red pairs of its separate evaluation cohort.

Attention mass measures the percentage of attention assigned to a token group $\mathcal G$ at VL layer $\ell$ (\site{2}):
\begin{equation}
 A_\ell(\mathcal G)=\frac{100}{NHQ}
 \sum_{n=1}^{N}\sum_{h=1}^{H}\sum_{q=1}^{Q}\sum_{k\in\mathcal G}
 \alpha^{(\ell)}_{n,h,q,k},
 \label{eq:attention_mass}
\end{equation}
where $\alpha$ is softmax attention, $N$ samples, $H$ heads, and $Q$ queries. We evaluate all language tokens and the color/order subset.

\subsection{Controlled Counterfactual Separation}
For each snapshot, the counterfactual changes only the instruction: RGB requires green next, while RBG requires blue. Camera observations, robot state, object configuration, and stochastic initialization remain fixed. Let $h_m^{\mathrm{RGB}}$ and $h_m^{\mathrm{RBG}}$ be the flattened action-query representations at a given DiT block (\site{4}). Their mean relative separation across $M$ snapshots is
\begin{equation}
 D_{\mathrm{rel}}=\frac1M\sum_{m=1}^{M}
 \frac{\|h_m^{\mathrm{RGB}}-h_m^{\mathrm{RBG}}\|_2}
 {\max(\|h_m^{\mathrm{RGB}}\|_2,\|h_m^{\mathrm{RBG}}\|_2,\epsilon)},
 \label{eq:relative_l2}
\end{equation}
with $\epsilon=10^{-12}$. This measures geometric separation, not information content. We retain the numerator $D_{\mathrm{abs}}$ and denominator $N_h$ to distinguish absolute distance from hidden-state magnitude. The primary analysis uses the first $K=16$ of the 40 action queries at the final flow-matching step, $t=3$, closest to decoded action. Endpoint controls at $B_0,B_{31}$ repeat the analysis for $t=0,1,2,3$ on the same cohort. Block increments $\Delta D_{\mathrm{rel}}(B_i)=D_{\mathrm{rel}}(B_i)-D_{\mathrm{rel}}(B_{i-1})$ are grouped by attention operation.

For each VL-adapter block (\site{3}), we instead mean-pool all valid sequence positions before computing relative separation. An image-only control pools positions selected by the image mask, excluding text. Adapter and DiT spaces are heterogeneous: their magnitudes are not differenced or placed on one depth axis. We compare within-space success--failure gaps and CIs to assess failure-associated separation within each representation space. Because updated action tokens are re-encoded at every flow-matching step, $B_0$ at $t=3$ already reflects three DiT passes; a gap there is therefore located within the action-token pathway without assignment to a block, operation, or first-arising stage.

Checkpoint changes use $\Delta_{\mathrm{SAM}}D_{\mathrm{rel}}(B_i)=D_{\mathrm{rel}}^{\mathrm{SAM}}(B_i)-D_{\mathrm{rel}}^{\mathrm{Base}}(B_i)$. The SAM-specific block increment is
\begin{equation}
\begin{split}
\Delta\Delta D_{\mathrm{rel}}(B_i)
={}&\Delta D_{\mathrm{rel}}^{\mathrm{SAM}}(B_i)
-\Delta D_{\mathrm{rel}}^{\mathrm{Base}}(B_i).
\end{split}
\label{eq:sam_increment}
\end{equation}
As a layout-geometry control, per-layout $D_{\mathrm{rel}}$ at $B_0$ and $B_{31}$ is correlated (Pearson and Spearman, layout-bootstrap CIs) with the three-dimensional green--blue cube center distance recorded in each post-red snapshot. Depth-curve CIs use 20,000 layout bootstrap resamples preserving complete block trajectories (Fig.~\ref{fig:paired_drel}); endpoint success--failure gaps independently resample within baseline cohorts, while checkpoint changes use paired layout resampling.

\subsection{Direct Cross-Modal Attention and Activation Patching}
We use attention routing to describe which image and text tokens receive attention from action queries. For Fig.~\ref{fig:cross_image}, direct image attention is averaged over heads, the first $K=16$ action queries, and the four denoising steps; the external-camera $8\times8$ maps at $B_{10},B_{26}$ (\site{5}) are bilinearly upsampled and shown on one linear scale (maximum 0.218, L06 at $B_{26}$) with identical colors, opacity, and dimming, preserving peak-strength differences. A camera's share of image attention is the sum over its 64 tokens. Text routing at $B_8,B_{28}$ is min--max normalized within each instruction span. These maps are representative routing evidence, not causal attribution.

Activation patching~\cite{zhang2024patching} (\site{6}) replaces RGB (recipient) states with their RBG (donor) counterparts. We replace the first 16 action-token output vectors at $B_{10},B_{18},B_{22},B_{26}$, or jointly at $B_{18}+B_{22}$ and $B_{18}+B_{22}+B_{26}$, at each of the four denoising steps. The prepended state token and remaining action tokens are not directly replaced. For flattened first-16 decoded actions $a_m^R,a_m^D,a_m^P$, let $d_m=a_m^D-a_m^R$ and $\Delta a_m=a_m^P-a_m^R$. Donor-directed leverage is
\begin{equation}
 E_{\mathrm{patch}}=\frac1M\sum_{m=1}^{M}
 \frac{\Delta a_m^\top d_m}{\max(\|d_m\|_2^2,\epsilon)}.
 \label{eq:epatch}
\end{equation}
Positive values indicate movement toward the donor action under this metric, negative values movement away. Replacement tests sufficiency, not necessity or unique localization, and later replacements can overwrite earlier ones.

\section{Results}
We first establish the behavioral failure pattern, then examine representations along the numbered sites in Fig.~\ref{fig:architecture}. These analyses primarily use AdamW 10k; the AdamW 20k replication uses a separate 20-state cohort. We then compare SAM with AdamW in behavior and representation geometry.

\subsection{Held-Out RGB Trails the Seen-Ordering Mean}
Under AdamW, held-out RGB trails the seen-ordering mean at both budgets (Table~\ref{tab:per_sequence_success}): 61.0\% versus 78.2\% at 10k (17.2-point gap) and 65\% versus 86.6\% at 20k (21.6-point gap). The within-checkpoint red-first contrast between RBG and RGB, which share their first placement, is 17 pp [3,32] (10k; exact paired layout-level $p=0.0547$) and 16 pp (20k) on familiar layouts and 25 pp [7,42] on disjoint layouts (AdamW 10k; Sec.~\ref{sec:unseen_layout_control}).

\subsection{Held-Out Failures Are Behaviorally Heterogeneous}
Table~\ref{tab:failure_composition} applies the same criteria to 100 familiar-layout RGB rollouts per checkpoint. Under AdamW 10k, observed or inferred wrong-second targets account for 15 of 39 failures; other failures involve execution, other stages, or unresolved selection. Observed wrong-second entries account for 20 of 35 AdamW 20k failures and 4 of 19 SAM failures.

\begin{table}[t]
\centering\scriptsize
\setlength{\tabcolsep}{2pt}
\caption{Failure-type composition of the 100 familiar-layout RGB rollouts per checkpoint (Sec.~III-D criteria; \%, layout-cluster bootstrap 95\% CIs). $^{\ddagger}$Four AdamW 20k rollouts in which another cube entered before red are not first-stage failures under the red-entered criterion.}
\label{tab:failure_composition}
\floatgap
\begin{tabular*}{\columnwidth}{@{\extracolsep{\fill}}lccc@{}}
\toprule
Category & AdamW 10k & AdamW 20k$^{\ddagger}$ & SAM 10k \\
\midrule
Success & 61 [46,76] & 65 [51,79] & 81 [67,93] \\
First-stage failure & 9 [1,19] & 3 [0,6] & 2 [0,5] \\
Wrong second, observed entry & 13 [3,25] & 20 [8,35] & 4 [0,11] \\
Wrong second, inferred & 2 [0,5] & 3 [0,6] & 2 [0,5] \\
Second target unresolved & 1 [0,3] & 0 [0,0] & 1 [0,3] \\
Correct second, execution failure & 7 [2,13] & 4 [0,9] & 5 [0,12] \\
Later-stage failure & 7 [2,14] & 5 [0,14] & 5 [0,11] \\
\bottomrule
\end{tabular*}
\end{table}

\subsection{Target Decodability across Representations}\label{sec:probe_results}
At both the Qwen backbone and final VL-adapter layer (\site{1}), the seen-task probe predicts the intended target in 19/19 RGB and 14/19 RBG snapshots, and green for every post-red RGB snapshot at every analyzed stage, including all five strict wrong-blue trajectories. Its errors fall on the RBG/blue branch, so it establishes availability of green-related information in RGB failures, not an unbiased current-target decoder.

The DiT probe tests transfer from seen after-green/after-blue states to post-red states. Seen-state validation accuracy is 87.5\% at $B_0$ and 97.5\% at $B_{31}$ (per-class recall 90/80/90\% and 100/95/95\%). Post-red RGB green recall is 11/19 and 8/19, RBG blue recall 10/19 and 17/19; within success/failure cohorts, RGB recall is 9/12 and 2/7 at $B_0$ versus 6/12 and 2/7 at $B_{31}$. Cross-prefix transfer is thus asymmetric rather than a uniform depth-wise decline, and distinct from geometric separation. On the AdamW 20k cohort (Sec.~III-D), the same protocol gives 83.8\%/96.3\% seen validation but post-red RGB green recall of only 5/20 and 7/20 (RBG blue 12/20 at both): weak transfer again favors blue and is not specific to the final block.

At the final VL self-attention layer (\site{2}), color/order attention is 3.090\% for RGB versus 3.035\% for the seen mean (total language attention 12.78\% versus 9.59\%), so RGB failure is not accompanied by a global loss of language routing.

\subsection{VL-Adapter Separation Shows No Detectable Failure-Associated Deficit}
\label{sec:adapter_results}
At every VL-adapter layer (\site{3}), mean separation is slightly larger in the failure cohort. Success$-$failure gaps range from $-0.003$ to $-0.032$ for all-token pooling and $-0.003$ to $-0.030$ for image-only pooling, with every 95\% layout-bootstrap CI including zero (layer 3: $[-0.101,0.033]$ and $[-0.107,0.042]$). The image-only control shows that changed color/ordinal text-token positions alone do not create this pattern.

The ordering reverses in the earliest analyzed DiT action-query representation (\site{4}, $t=3$): the AdamW success--failure gap is $+0.0685$ $[+0.0168,+0.1271]$. A failure-associated separation deficit is therefore detectable downstream of site \site{3}, throughout the DiT action-token pathway. However, $B_0$ at $t=3$ already reflects three denoising passes (Sec.~III-F), while its gap is near zero at $t=0$ (Sec.~\ref{sec:dit_results}). The analysis neither identifies where the deficit first arises nor implicates or excludes the adapter-to-DiT boundary.

\begin{figure*}[t]
\centering
\includegraphics[width=\figscale\textwidth]{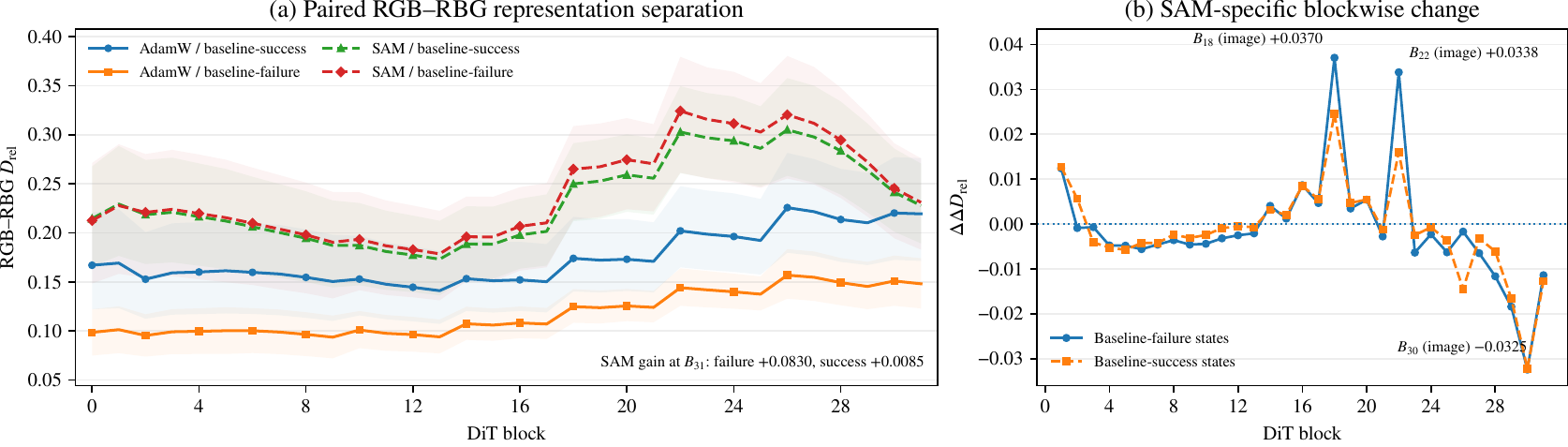}
\floatgapwide
\caption{\textbf{Instruction-conditioned action separation and its SAM-specific increments.} (a) $D_{\mathrm{rel}}$ at $t=3$, $K=16$, on identical post-red observations, grouped by original AdamW 10k success ($n=12$) or failure ($n=7$); shading is layout-bootstrap 95\% CI. The AdamW gap persists from $B_0$ to $B_{31}$; SAM curves largely overlap. (b) $\Delta\Delta D_{\mathrm{rel}}$: failure-cohort excursions are larger at $B_{18}$ (+0.0370) and $B_{22}$ (+0.0338).}
\label{fig:paired_drel}
\end{figure*}

\subsection{A Separation Deficit Persists Through DiT Action Generation}
\label{sec:dit_results}
The AdamW 10k separation gap in Fig.~\ref{fig:paired_drel}(a) (site \site{4}) persists from the earliest analyzed DiT representation (Sec.~\ref{sec:adapter_results}) to $B_{31}$: $+0.0713$ $[+0.0163,+0.1338]$, success/failure means 0.2194/0.1481; action generation adds separation to both groups without eliminating the deficit.

\paragraph{Layout-geometry control}
Because separation is measured on action queries, $D_{\mathrm{rel}}$ could track the distance between the candidate targets. Across the 19 primary snapshots, the green--blue distance recorded in the post-red state is only weakly associated with $D_{\mathrm{rel}}$ (Pearson $r=0.03$ $[-0.39,0.55]$ at $B_0$, $0.06$ $[-0.35,0.59]$ at $B_{31}$), and mean distances are similar for the success and failure cohorts (0.160 versus 0.151 m); this does not exclude spatial confounding at $n=19$ but gives no support to a distance-only account of the 10k gap.

\paragraph{AdamW 20k replication}
The 20-state AdamW 20k cohort has same-direction gaps at $B_0$ and $B_{31}$ ($+0.053$ and $+0.078$, CIs including zero), but $D_{\mathrm{rel}}$ correlates strongly with green--blue distance ($r\approx0.73$ at both blocks), and the four failure layouts have shorter mean distances than successful layouts (0.119 versus 0.159 m). This is not an independent replication of the outcome-associated deficit because the gap remains inseparable from scene geometry.

\paragraph{Denoising-step sensitivity}
The $B_{31}$ AdamW gap is positive with CIs excluding zero at all four flow-matching steps ($+0.0426$, $+0.0460$, $+0.0526$, $+0.0713$ for $t=0,\ldots,3$), while all SAM intervals span zero; at $B_0$ the gap is near zero at $t=0$, where both branches share the initial noise, and positive from $t=1$. At $t=3$, image cross-attention contributes the largest block increment in both cohorts ($+0.0119$ success, $+0.0097$ failure); this shared pattern is not a failure-specific discriminator.

\begin{figure*}[t]
\centering
\includegraphics[width=\figscale\textwidth]{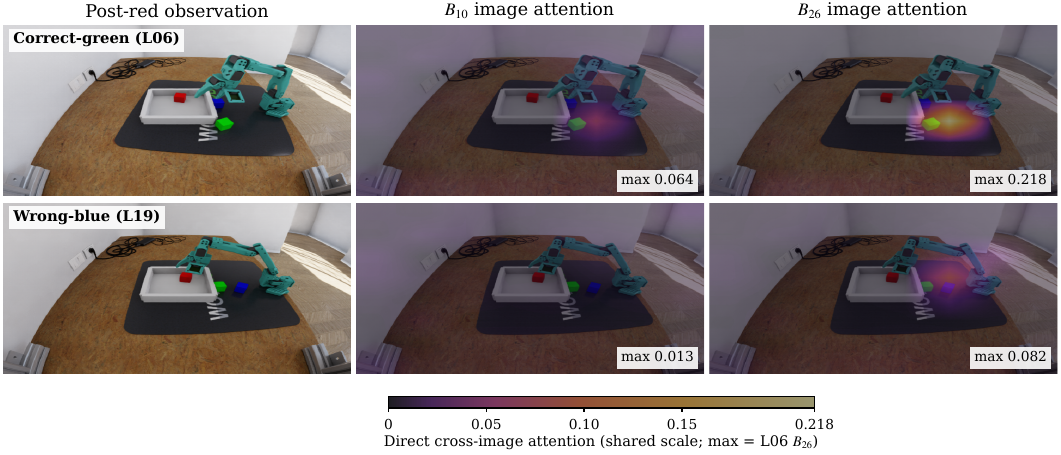}
\floatgapwide
\caption{\textbf{Representative direct visual routing during DiT action generation.} Rows: correct-green L06 and wrong-blue L19; columns: post-red observation and image attention at $B_{10},B_{26}$, averaged over denoising steps and upsampled from $8\times8$ token grids. All maps share one linear color scale (maximum 0.218, L06 at $B_{26}$). At $B_{26}$ the external-camera attention shares are nearly equal (39.74\% and 39.79\%), and the wrong-blue map includes the region of the subsequently selected blue cube.}

\label{fig:cross_image}
\end{figure*}

\begin{figure}[t]
\centering
\includegraphics[width=\figscale\columnwidth]{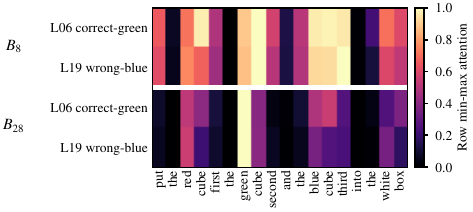}
\floatgap
\caption{\textbf{Representative language routing.} Direct attention from $K=16$ action queries at $B_8$ and $B_{28}$ for correct-green L06 and wrong-blue L19 (one row per block and trajectory), min--max normalized per row over the 18 word tokens (punctuation not retained), so colors show relative preference, not absolute mass or causal attribution. Both late maps retain a relative preference for ``green.''}
\label{fig:cross_text}
\end{figure}

\subsection{Representative Routing Shows a Cross-Modal Dissociation}
In wrong-blue L19, late text routing (\site{5}) retains a relative preference for ``green,'' while late image routing includes the subsequently selected blue region (Figs.~\ref{fig:cross_image}--\ref{fig:cross_text}); external-camera attention shares are nearly equal at $B_{26}$ in the two examples and differ at $B_{10}$ (39.61\% versus 33.60\%). This two-example dissociation is consistent with an action-conditioning account but is not a cohort-level statistic.

\begin{figure*}[t]
\centering
\includegraphics[width=\figscale\textwidth]{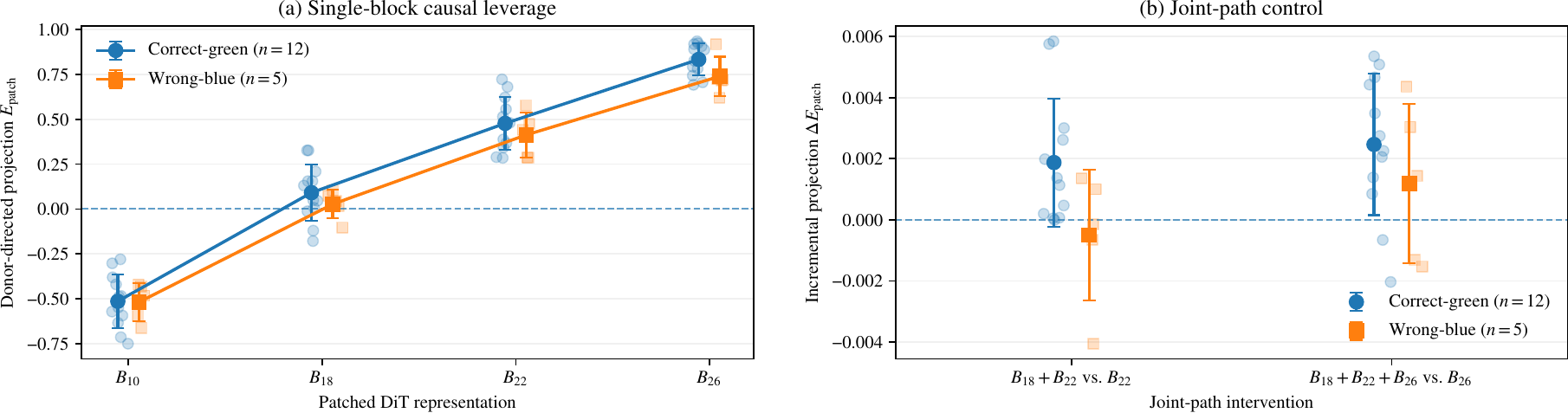}
\floatgapwide
\caption{\textbf{Depth-dependent action leverage.} (a) Donor-directed projection after replacing the first 16 action-token output vectors at all four denoising steps for correct-green ($n=12$) and wrong-blue ($n=5$); two unresolved layouts are excluded. Points are layouts; error bars are mean $\pm$ SD. (b) Joint replacements add little beyond the deepest single replacement.}
\label{fig:activation_patching}
\end{figure*}

\subsection{Activation Patching Reveals Depth-Dependent Causal Leverage}
Donor-directed displacement grows with depth in both strict cohorts (Fig.~\ref{fig:activation_patching}, site \site{6}): mean $E_{\mathrm{patch}}$ for correct-green / wrong-blue trajectories is $-0.5133/-0.5176$ at $B_{10}$, $+0.0912/+0.0273$ at $B_{18}$, $+0.4768/+0.4130$ at $B_{22}$, and $+0.8331/+0.7392$ at $B_{26}$. This shared pattern locates output leverage, not a failure-specific representation. The $B_{10}$ projection is negative in all 17 strict-cohort layouts and leave-one-layout-out analyses (means $-0.529$ to $-0.500$); its mechanism remains unresolved. Joint replacements add $|\Delta E_{\mathrm{patch}}|<0.003$ over the deepest single replacement. Because later replacements overwrite earlier ones, panel (b) controls the intervention path.

\subsection{SAM Improves Held-Out RGB Execution}
\label{sec:sam_results}

\begin{table}[t]
\centering\scriptsize
\setlength{\tabcolsep}{2pt}
\caption{Familiar-layout success (\%) at 10k or 20k optimizer updates. Each cell gives an estimate and layout-cluster bootstrap 95\% CI; 20 layouts $\times5$ repeats per ordering.}
\label{tab:per_sequence_success}
\floatgap
\begin{tabular*}{\columnwidth}{@{\extracolsep{\fill}}lccc@{}}
\toprule
Ordering & AdamW 10k & AdamW 20k & SAM 10k \\
\midrule
GRB & 87 [79,94] & 90 [83,96] & 88 [81,94] \\
BGR & 70 [60,80] & 90 [82,97] & 88 [81,94] \\
BRG & 72 [62,81] & 85 [73,94] & 82 [72,91] \\
GBR & 84 [73,93] & 87 [79,94] & 86 [76,94] \\
RBG & 78 [65,89] & 81 [70,91] & 90 [82,97] \\
Seen mean & 78.2 [72.2,84.0] & 86.6 [81.4,91.2] & 86.8 [81.6,91.2] \\
RGB (OOD) & 61 [46,76] & 65 [51,79] & 81 [67,93] \\
\bottomrule
\end{tabular*}
\end{table}

With the number of forward--backward passes matched (Table~\ref{tab:per_sequence_success}), AdamW 20k and SAM 10k reach similar seen mean success, 86.6\% and 86.8\%, but RGB success is 65\% versus 81\% (+16 pp for SAM [2,30], exact paired $p=0.058$), and directly observed wrong-blue second entries occur in 20/100 versus 4/100 RGB rollouts ($-16$ pp [$-27$,$-7$], $p=0.0078$). The seen--RGB gap is +21.6 pp for AdamW 20k and +5.8 pp [$-5.8$,18.8] for SAM.

Relative to AdamW 10k (matched optimizer updates), SAM adds +8.6 pp seen success ([3.6,13.8], $p=0.00595$) and +20 pp RGB success ([7,34], $p=0.015$), and the seen--RGB gap falls from +17.2 pp [1.4,33.4] to +5.8 pp (reduction 11.4 [$-3.8$,27.0], $p=0.18$); only BGR survives Holm correction ($p_{\mathrm{Holm}}=0.0458$). Observed wrong-blue entries are 13 and 20 per 100 rollouts for the two AdamW runs (+7 pp [$-3$,18], exact paired $p=0.3125$). Doubling the AdamW budget thus improves seen execution without closing the RGB gap; SAM's gap at the same number of passes includes zero in its CI, but its reduction relative to AdamW 10k is not resolved ($p=0.18$).

\paragraph{$\rho$ sensitivity}
The SAM 10k runs at $\rho=0.01$ and $\rho=0.1$ (Sec.~III-B) reach seen means of 90.6\% and 89.6\% but RGB success of 74\% and 75\% (RBG 88\% and 91\%; wrong-blue 13/100 and 7/100), so their red-first asymmetries ($+14$, $+16$ pp) and seen--RGB gaps ($+16.6$, $+14.6$ pp) are comparable in magnitude to those under AdamW. $\rho=0.05$ exceeds them on RGB by $+7$ pp [$-5$,19] and $+6$ pp [$-4$,16] (exact paired $p=0.36$, $0.37$); against AdamW 20k, RGB is $+16$ pp [2,30] ($p=0.058$) at $\rho=0.05$ and $+9$ pp [$-4$,23] ($p=0.27$) and $+10$ pp [$-5$,26] ($p=0.28$) at $\rho=0.01$/$0.1$, with wrong-blue reductions of $-7$ pp [$-17$,1] and $-13$ pp [$-23$,$-5$]. All three SAM runs exceed AdamW 20k in the same direction; only the $\rho=0.05$ difference has a CI excluding zero, and one run per $\rho$ does not separate hyperparameter sensitivity from training-run variability.

\paragraph{Real-robot color-order evaluation}
Table~\ref{tab:real_robot} summarizes the physical-robot trials: AdamW 20k followed the instructed order in 18/30 RGB and 26/30 RBG trials (red-first contrast 27 pp, Fisher $p=0.039$) and SAM ($\rho=0.05$) in 24/30 and 29/30 (17 pp, $p=0.10$); RGB differs by +20 pp between checkpoints ($p=0.16$) and RBG by +10 pp ($p=0.35$). Blue placed second after a correct red placement is the most frequent RGB failure for both (7/30 versus 3/30, $p=0.30$); every RBG failure placed a wrong cube first. The red-first asymmetry and its dominant failure mode therefore reproduce on hardware with policies trained only on real demonstrations, and the SAM pair moves in the same direction as in simulation. These rates measure color order with grasp-failure restarts, not end-to-end execution.

\begin{table}[t]
\centering\scriptsize
\setlength{\tabcolsep}{2.5pt}
\caption{Real-robot color-order evaluation on the physical SO-101 (30 unpaired trials per cell; separately fine-tuned policies; SAM at $\rho=0.05$). Success: instructed order followed, with Wilson 95\% CIs. G-first/B-first: wrong cube placed first, then red; wrong 2nd: blue placed after a correct red placement (RGB).}
\label{tab:real_robot}
\floatgap
\begin{tabular*}{\columnwidth}{@{\extracolsep{\fill}}llcccc@{}}
\toprule
Checkpoint & Order & Success & G-first & B-first & Wrong 2nd \\
\midrule
AdamW 20k & RGB (OOD) & 18/30 (60.0 [42.3,75.4]) & 4 & 1 & 7 \\
AdamW 20k & RBG (seen) & 26/30 (86.7 [70.3,94.7]) & 2 & 2 & 0 \\
SAM ($\rho=0.05$) & RGB (OOD) & 24/30 (80.0 [62.7,90.5]) & 2 & 1 & 3 \\
SAM ($\rho=0.05$) & RBG (seen) & 29/30 (96.7 [83.3,99.4]) & 0 & 1 & 0 \\
\bottomrule
\end{tabular*}
\end{table}

\subsection{The Red-First Contrast Persists on Disjoint Layouts}
\label{sec:unseen_layout_control}

On 20 disjoint layouts, RBG exceeds RGB by 25 pp [7,42] under AdamW 10k (exact $p=0.0182$) but by 1 pp [$-10,12$] under SAM ($p=1.0$), a 24-pp reduction [6,42] (paired $p=0.0287$); familiar contrasts are $+17$ pp [3,32] and $+9$ pp [$-3$,22] (reduction 8 pp [$-9$,26], $p=0.47$). The direct disjoint RGB gain is $+34$ pp [$+20,+48$] ($p=0.00058$), while the $+10$-pp RBG change includes zero ($p=0.112$).

Table~\ref{tab:wrong_second_entries} reports wrong-second entries: SAM reduces disjoint wrong-blue entries by 15 pp [$-24,-7$] ($p=0.0078$) and familiar entries from 13\% to 4\% ($-9$ pp [$-18$,$-1$], $p=0.125$). The AdamW asymmetry persists off demonstration layouts and is attenuated by SAM; representation analyses remain based on familiar layouts.

\begin{table}[t]
\centering\scriptsize
\setlength{\tabcolsep}{2pt}
\caption{Disjoint-layout success (\%) and SAM--AdamW 10k change (pp), with layout-cluster bootstrap 95\% CIs; 20 layouts $\times5$ repeats.}
\label{tab:unseen_layout_success}
\floatgap
\begin{tabular*}{\columnwidth}{@{\extracolsep{\fill}}lccc@{}}
\toprule
Ordering & AdamW 10k & SAM 10k & $\Delta$ \\
\midrule
RGB (OOD) & 49 [35,64] & 83 [74,91] & +34 [+20,+48] \\
RBG (seen) & 74 [62,85] & 84 [75,93] & +10 [0,+20] \\
\bottomrule
\end{tabular*}
\end{table}

\begin{table}[t]
\centering
\scriptsize
\caption{Directly observed wrong-second entries for the red-first RGB/RBG
comparison. Layout-cluster bootstrap 95\% CIs in brackets. $\dagger$ Familiar RGB AdamW--SAM comparison: exact paired layout-level $p=0.125$.}
\label{tab:wrong_second_entries}
\floatgap
\begin{tabular}{llcc}
\toprule
\textbf{Layouts} & \textbf{Checkpoint} & \textbf{RGB: blue} & \textbf{RBG: green} \\
\midrule
Familiar & AdamW 10k & 13\% [3,25]$^{\dagger}$ & 1\% [0,3] \\
Familiar & SAM 10k & 4\% [0,11]$^{\dagger}$ & 0\% [0,0] \\
Disjoint & AdamW 10k & 17\% [8,26] & 1\% [0,3] \\
Disjoint & SAM 10k & 2\% [0,6] & 1\% [0,3] \\
\bottomrule
\end{tabular}
\end{table}

\paragraph{Success gains versus wrong-blue reductions}
On familiar layouts, the AdamW 20k--SAM difference is largest for wrong-second entries ($-16$ pp [$-27$,$-7$]); other categories change by at most 1 pp, and relative to AdamW 10k first-stage and wrong-second failures change by $-7$ and $-9$ pp. These changes are not a decomposition of the success gain, and SAM's reduction of the red-first asymmetry (resolved on disjoint, not on familiar layouts) does not establish how it produces the improvement.

\subsection{SAM Reduces the Observed Cohort Gap in Action-Token Separation}
On the same 19 post-red observations, with initial noise and AdamW 10k-defined cohort labels fixed, the estimated separation gaps between cohorts are close to zero under SAM while hidden-state scale also changes: paired $D_{\mathrm{rel}}$ changes for failure/success are $+0.1140/+0.0474$ at $B_0$ and $+0.0830/+0.0085$ at $B_{31}$, with SAM gaps $+0.0019$ [$-0.0757,+0.0840$] and $-0.0032$ [$-0.0634,+0.0603$]. The $B_{29}\!\rightarrow\!B_{30}$ relative decrease for failures coincides with increases in absolute distance ($+32.1$) and denominator ($+403.9$); AdamW success/failure absolute distances are 143.1/84.4 at $B_0$ and 864.9/586.2 at $B_{31}$ with nearly equal denominators, whereas SAM's are outcome-balanced (106.7/105.7; 716.9/728.7) with denominators 42\% and 20\% smaller. Cross-checkpoint changes therefore mix branch distance and hidden-state scale.

\section{Discussion}
Behavioral tests show a red-first asymmetry on familiar and disjoint layouts. Familiar-layout analyses find decodable upstream targets, a geometric separation deficit through the DiT action-token pathway, and stronger action leverage at later states. At matched forward--backward passes, SAM at $\rho=0.05$ matches the AdamW 20k seen mean, improves RGB without RGB demonstrations, reduces the disjoint-layout asymmetry, and reduces the baseline-cohort separation gap; the separately trained real-robot pair shows the same direction (60\% versus 80\% RGB color order) and the same most frequent RGB failure, although the between-checkpoint difference is inconclusive at 30 trials per condition.

These data do not show that RBG demonstrations caused the asymmetry, that SAM corrected target selection, or that geometric changes mediate the gain. The observed separation deficit does not by itself imply a loss of target information and is only weakly associated with inter-cube distance in the 10k cohort, while the 20k replication is confounded by scene geometry. The SAM result is an empirical checkpoint comparison---RGB gains in the same direction at all three $\rho$ values (74--81\%), largest at $\rho=0.05$---and does not show that reduced sharpness causes the improvement or that longer AdamW training could not close the gap.

\section{Limitations}
\label{sec:limitations}
This report covers one model, three objects, one simulation training run per optimizer--budget setting (the real-robot policies are separate runs on real demonstrations), and RGB held out from five orderings. CIs and $p$-values quantify evaluation variance for fixed checkpoints, not run-to-run training variance; AdamW budgets beyond 20k and leave-one-permutation-out studies are untested, and mechanistic analyses reuse demonstration-layout support while disjoint layouts test only RGB/RBG in one workspace with one instruction template. Small snapshot cohorts limit precision, and separation can covary with inter-cube distance, especially in the 20k cohort; probe asymmetry and cross-prefix transfer limit decodability claims. Routing panels are representative; image-token projection, alternative-donor, reverse-direction, per-joint scaling, and context-token replacement controls were not performed. The $\rho=0.01$ and $0.1$ runs reach 74--75\% RGB with gains over AdamW 20k whose CIs include zero, and the $\rho=0.05$ gain has an exact paired $p$ of 0.058; one run per $\rho$ cannot separate hyperparameter sensitivity from training-run variability. The real-robot evaluation uses 30 unpaired trials per ordering, restarts grasp failures, and measures color order rather than end-to-end success.

\section{Conclusion}
The RGB holdout reveals a red-first asymmetry on familiar and disjoint layouts; familiar-layout analyses find decodable upstream targets and a downstream geometric separation deficit. At matched forward--backward passes, SAM at $\rho=0.05$ matches AdamW 20k seen performance, improves RGB without RGB demonstrations, reduces the disjoint-layout asymmetry, and reduces the baseline-cohort separation gap; the red-first asymmetry, its blue-second failure mode, and the direction of the SAM difference reproduce on a physical SO-101 with separately trained policies. Together, these findings highlight the distinction between upstream target decodability and downstream action selection in this sequencing setting.


{\footnotesize
\begin{thebibliography}{99}
\setlength{\itemsep}{0pt}
\setlength{\parskip}{0pt}

\bibitem{brohan2023rt2}A. Brohan et al., ``RT-2: Vision-Language-Action Models Transfer Web Knowledge to Robotic Control,''
in \emph{CoRL}, 2023.


\bibitem{kim2024openvla}M. J. Kim et al., ``OpenVLA: An Open-Source Vision-Language-Action Model,''
in \emph{CoRL}, 2024.

\bibitem{octo2024}D. Ghosh et al., ``Octo: An Open-Source Generalist Robot Policy,''
in \emph{RSS}, 2024.

\bibitem{nvidia2025gr00tn1}J. Bjorck et al., ``GR00T N1: An Open Foundation Model for Generalist Humanoid Robots,'' arXiv:2503.14734, 2025.

\bibitem{jiang2022vima}Y. Jiang et al., ``VIMA: General Robot Manipulation with Multimodal Prompts,''
in \emph{ICML}, 2023.

\bibitem{peng2026acvla}X. Peng et al., ``AC-VLA: Robust Out-of-Distribution Action Execution via Compositional Learning,'' arXiv:2607.15714, 2026.


\bibitem{dehaan2019causal}P. de Haan, D. Jayaraman, and S. Levine,
``Causal Confusion in Imitation Learning,''
in \emph{NeurIPS}, 2019.

\bibitem{geirhos2020shortcut}R. Geirhos et al.,
``Shortcut Learning in Deep Neural Networks,''
\emph{Nat. Mach. Intell.}, vol. 2, 2020.

\bibitem{nvidia2026n17}NVIDIA,
``Isaac GR00T N1.7,'' 2026. [Online]. Available: \url{https://github.com/NVIDIA/Isaac-GR00T}

\bibitem{foret2021sam}P. Foret, A. Kleiner, H. Mobahi, and B. Neyshabur, ``Sharpness-Aware Minimization for Efficiently Improving Generalization,'' in \emph{ICLR}, 2021.

\bibitem{openx2023}Open X-Embodiment Collaboration et al., ``Open X-Embodiment: Robotic Learning Datasets and RT-X Models,''
in \emph{ICRA}, 2024.


\bibitem{black2025pi0}K. Black et al.,
``$\pi_0$: A Vision-Language-Action Flow Model for General Robot Control,''
in \emph{RSS}, 2025.

\bibitem{chi2023diffusionpolicy}C. Chi et al., ``Diffusion Policy: Visuomotor Policy Learning via Action Diffusion,'' in \emph{RSS}, 2023.

\bibitem{lipman2023flowmatching}Y. Lipman et al., ``Flow Matching for Generative Modeling,'' in \emph{ICLR}, 2023.

\bibitem{peebles2023dit}W. Peebles and S. Xie,
``Scalable Diffusion Models with Transformers,''
in \emph{ICCV}, 2023.

\bibitem{lake2018systematicity}B. M. Lake and M. Baroni,
``Generalization without Systematicity: On the Compositional Skills of Sequence-to-Sequence Recurrent Networks,''
in \emph{ICML}, 2018.

\bibitem{hupkes2020compositionality}D. Hupkes, V. Dankers, M. Mul, and E. Bruni,
``Compositionality Decomposed: How Do Neural Networks Generalise?,''
\emph{JAIR}, vol. 67, 2020.

\bibitem{ruis2020gscan}L. Ruis et al.,
``A Benchmark for Systematic Generalization in Grounded Language Understanding,''
in \emph{NeurIPS}, 2020.

\bibitem{shridhar2022cliport}M. Shridhar, L. Manuelli, and D. Fox,
``CLIPort: What and Where Pathways for Robotic Manipulation,''
in \emph{CoRL}, 2022.

\bibitem{wang2023programport}R. Wang et al.,
``Programmatically Grounded, Compositionally Generalizable Robotic Manipulation,''
in \emph{ICLR}, 2023.

\bibitem{belinkov2022probing}Y. Belinkov, ``Probing Classifiers: Promises, Shortcomings, and Advances,'' \emph{Comput. Linguist.}, vol. 48, no. 1, 2022.

\bibitem{elazar2021amnesic}Y. Elazar et al.,
``Amnesic Probing: Behavioral Explanation with Amnesic Counterfactuals,''
\emph{TACL}, vol. 9, 2021.

\bibitem{jain2019attention}S. Jain and B. C. Wallace,
``Attention is not Explanation,''
in \emph{NAACL-HLT}, 2019.

\bibitem{meng2022rome}K. Meng et al., ``Locating and Editing Factual Associations in GPT,'' in \emph{NeurIPS}, 2022.

\bibitem{zhang2024patching}F. Zhang and N. Nanda,
``Towards Best Practices of Activation Patching in Language Models: Metrics and Methods,''
in \emph{ICLR}, 2024.

\bibitem{grant2026features}B. Grant, X. Zhao, and P. Wang, ``Not All Features Are Created Equal: A Mechanistic Study of Vision-Language-Action Models,''
in \emph{ICLR}, 2026.


\bibitem{keskar2017large}N. S. Keskar et al.,
``On Large-Batch Training for Deep Learning: Generalization Gap and Sharp Minima,''
in \emph{ICLR}, 2017.

\bibitem{kwon2021asam}J. Kwon et al.,
``ASAM: Adaptive Sharpness-Aware Minimization for Scale-Invariant Learning of Deep Neural Networks,''
in \emph{ICML}, 2021.

\bibitem{bahri2022sam}D. Bahri, H. Mobahi, and Y. Tay,
``Sharpness-Aware Minimization Improves Language Model Generalization,''
in \emph{ACL}, 2022.

\bibitem{mittal2023orbit}M. Mittal et al.,
``Orbit: A Unified Simulation Framework for Interactive Robot Learning Environments,''
\emph{IEEE Robot. Autom. Lett.}, vol. 8, no. 6, 2023.

\bibitem{loshchilov2019adamw}I. Loshchilov and F. Hutter,
``Decoupled Weight Decay Regularization,''
in \emph{ICLR}, 2019.

\end{thebibliography}
}
\end{document}